%% file: main.tex
\documentclass[a4paper, 10pt, conference]{ieeeconf}

\usepackage{FG2020}
\usepackage{graphicx}
\usepackage{float}
\usepackage{subfigure}

\usepackage{dblfloatfix}
\usepackage{cite}
\IEEEoverridecommandlockouts                          
\def\FGPaperID{114} % *** Enter the FG2020 Paper ID here

\title{\LARGE \bf
Efficient Human Pose Estimation by Maximizing Fusion and High-Level Spatial Attention
}

\author{\parbox{16cm}{\centering
    {\large Zhiyuan Ren$^1$, Yaohai Zhou$^2$, Yizhe Chen$^2$, Ruisong Zhou$^3$, Yayu Gao$^*$}\\
    }% <-this % stops a space
    \\
    a3295715438@gmail.com
    \\ Paper ID
    \FGPaperID \\
}

\begin{document}

\ifFGfinal
\thispagestyle{empty}
\pagestyle{empty}
\else
\author{Zhiyuan Ren$^1$, Yaohai Zhou$^2$, Yizhe Chen$^2$, Ruisong Zhou$^3$, Yayu Gao$^*$\\}
\pagestyle{plain}
\fi
\maketitle

%%%%%%%%%%%%%%%%%%%%%%%%%%%%%%%%%%%%%%%%%%%%%%%%%%%%%%%%%%%%%%%%%%%%%%%%%%%%%%%%
\input{abstract}
%%%%%%%%%%%%%%%%%%%%%%%%%%%%%%%%%%%%%%%%%%%%%%%%%%%%%%%%%%%%%%%%%%%%%%%%%%%%%%%%

%%%%%%%%%%%%%%%%%%%%%%%%%%%%%%%%%%%%%%%%%%%%%%%%%%%%%%%%%%%%%%%%%%%%%%%%%%%%%%%%
\input{introduction}

%%%%%%%%%%%%%%%%%%%%%%%%%%%%%%%%%%%%%%%%%%%%%%%%%%%%%%%%%%%%%%%%%%%%%%%%%%%%%%%%

%%%%%%%%%%%%%%%%%%%%%%%%%%%%%%%%%%%%%%%%%%%%%%%%%%%%%%%%%%%%%%%%%%%%%%%%%%%%%%%%
\input{related_work}
%%%%%%%%%%%%%%%%%%%%%%%%%%%%%%%%%%%%%%%%%%%%%%%%%%%%%%%%%%%%%%%%%%%%%%%%%%%%%%%%

%%%%%%%%%%%%%%%%%%%%%%%%%%%%%%%%%%%%%%%%%%%%%%%%%%%%%%%%%%%%%%%%%%%%%%%%%%%%%%%%
\input{method}
%%%%%%%%%%%%%%%%%%%%%%%%%%%%%%%%%%%%%%%%%%%%%%%%%%%%%%%%%%%%%%%%%%%%%%%%%%%%%%%%

%%%%%%%%%%%%%%%%%%%%%%%%%%%%%%%%%%%%%%%%%%%%%%%%%%%%%%%%%%%%%%%%%%%%%%%%%%%%%%%%
\input{experiment}
%%%%%%%%%%%%%%%%%%%%%%%%%%%%%%%%%%%%%%%%%%%%%%%%%%%%%%%%%%%%%%%%%%%%%%%%%%%%%%%%

%%%%%%%%%%%%%%%%%%%%%%%%%%%%%%%%%%%%%%%%%%%%%%%%%%%%%%%%%%%%%%%%%%%%%%%%%%%%%%%%
\input{conclusion}
%%%%%%%%%%%%%%%%%%%%%%%%%%%%%%%%%%%%%%%%%%%%%%%%%%%%%%%%%%%%%%%%%%%%%%%%%%%%%%%%

\pagebreak[4]

% \vspace{7cm}
%\bibliographystyle{plain}
\bibliographystyle{ieeetr}
\bibliography{reference}
\end{document}

%% file: abstract.tex
\begin{abstract}

In this paper, we propose an efficient human pose estimation network---SFM (slender fusion model) by fusing multi-level features and adding lightweight attention blocks---HSA (High-Level Spatial Attention). Many existing methods on efficient network have already taken feature fusion into consideration, which largely boosts the performance. However, its performance is far inferior to large network such as ResNet and HRNet due to its limited fusion operation in the network. Specifically, we expand the number of fusion operation by building bridges between two pyramid frameworks without adding layers. Meanwhile, to capture long-range dependency, we propose a lightweight attention block---HSA, which computes second-order attention map. In summary, SFM maximizes the number of feature fusion in a limited number of layers. HSA learns high precise spatial information by computing the attention of spatial attention map. With the help of SFM and HSA, our network is able to generate multi-level feature and extract precise global spatial information with little computing resource. Thus, our method achieve comparable or even better accuracy with less parameters and computational cost. Our SFM achieve 89.0 in PCKh@0.5, 42.0 in PCKh@0.1 on MPII validation set and 71.7 in AP, 90.7 in AP@0.5 on COCO validation with only 1.7G FLOPs and  1.5M parameters. The source code will be public soon.
 
\end{abstract}

%% file: introduction.tex
\section{Introduction}

Pose estimation is a fundamental task in computer vision, which mainly focuses on locating spatial coordinates of human joints in a 2D picture. Pose estimation task can boost the development of many novel fields such as human computer interaction games~\cite{en2010social}, auto driving~\cite{kothari2017pose} and so on. It is a non-trival task as the appearance of body joints varies dramatically due to flexible body structure, diverse style of clothes and complex background contexts. 

In previous study it was found that multi-scale feature fusion operation~\cite{qin2020multi, yang2020multi} and long-distance dependency capture can help address these questions. The former one refers to that the feature map will lose some important spatial information after convolution operations. So by adding some fusion connections between layers integrating multi-scale information properly can help neural network locate the position of keypoints from more information source. 
The latter one means computing the spatial dependency between human keypoints and this dependency can help judge the position of some difficult-identifying keypoints. For example, we can locate the location of a hidden elbow based on the location of wrist and shoulders.
\begin{figure}[htbp]
\centering
\subfigure[PCKh@0.5 versus Params]{
\includegraphics[width=3.8cm,height=3.8cm]{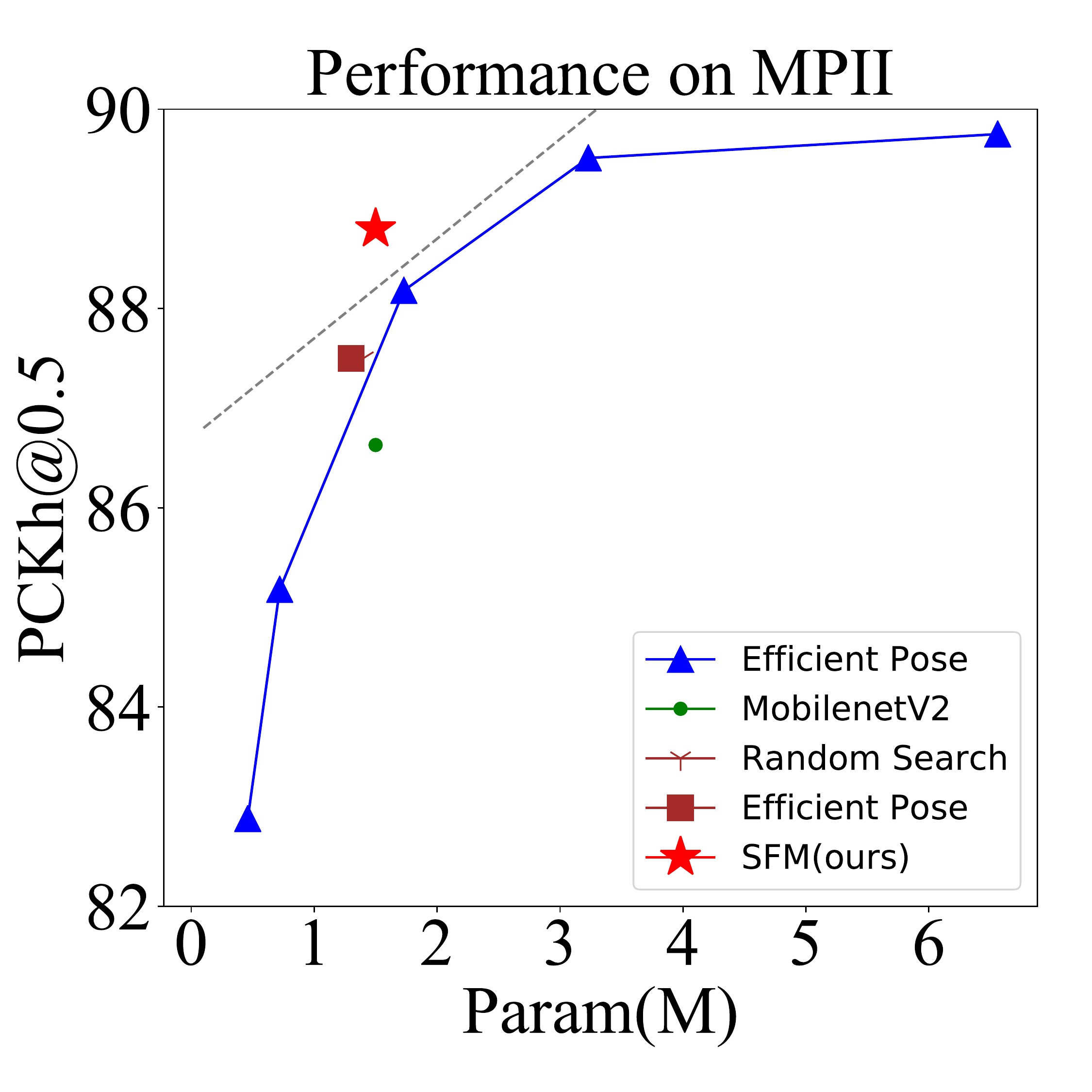}
}
\quad
\subfigure[PCKh@0.5 versus FLOPs]{
\includegraphics[width=3.8cm,height=3.8cm]{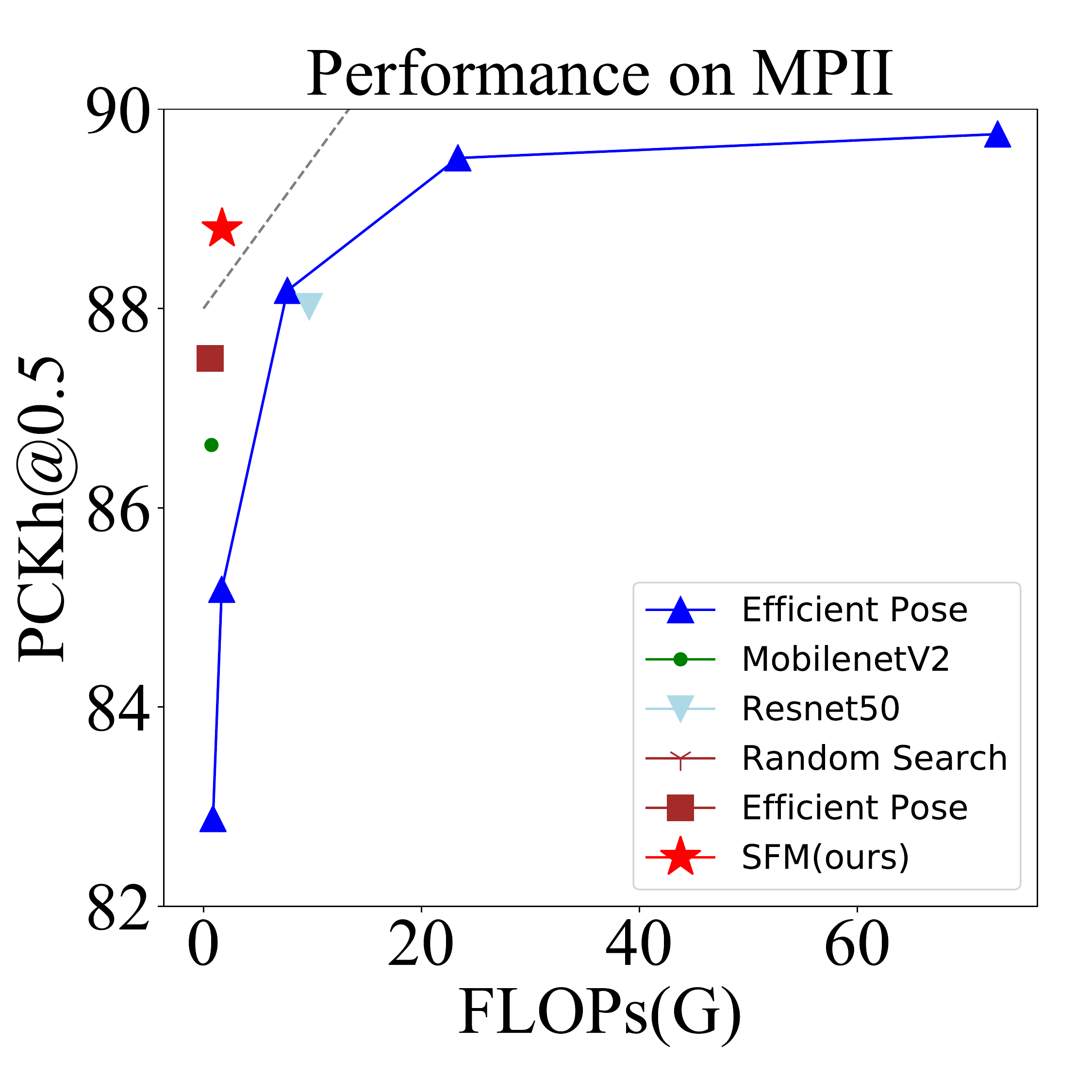}
}
\quad
\subfigure[AP versus Params]{
\includegraphics[width=3.8cm,height=3.8cm]{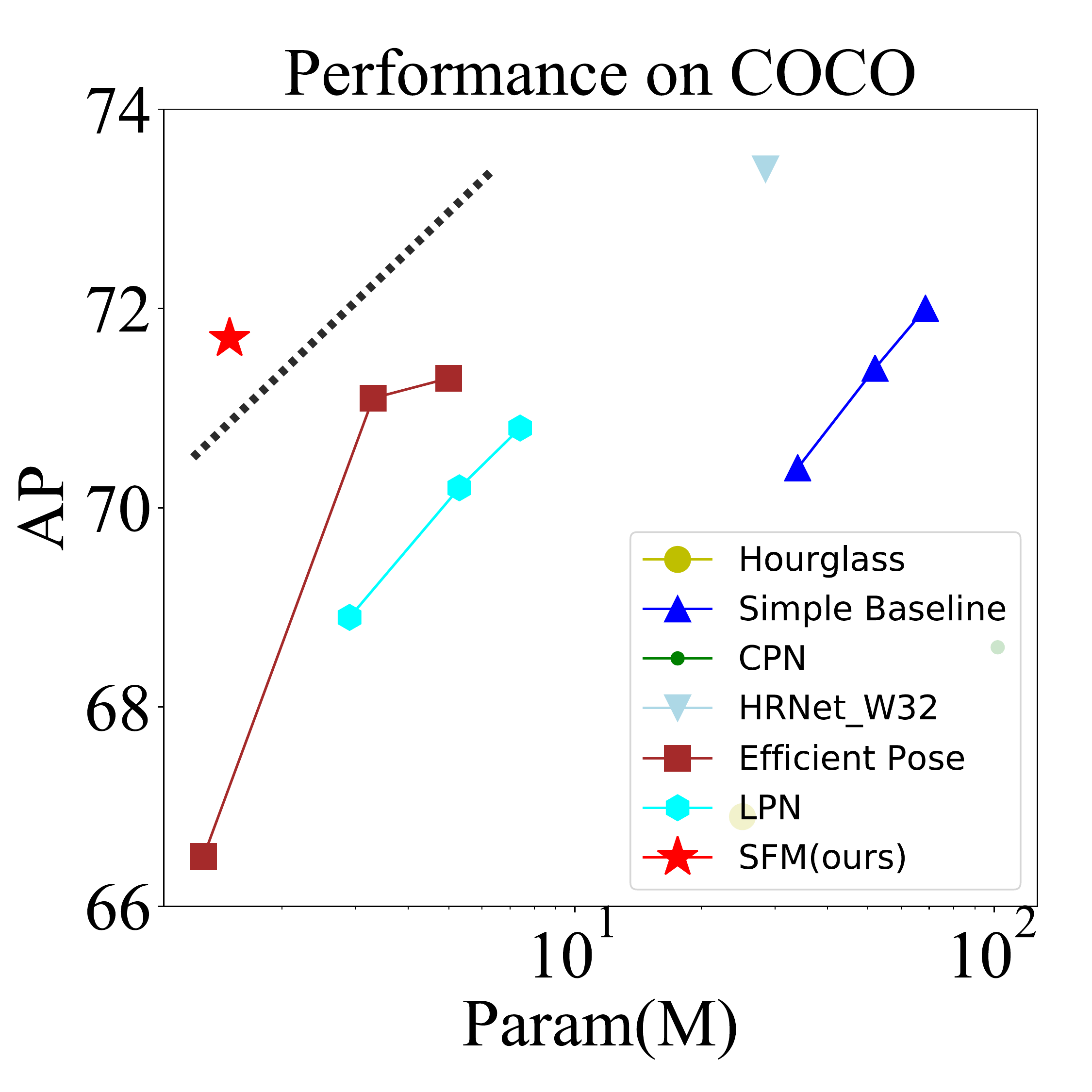}
}
\quad
\subfigure[AP versus FLOPs]{
\includegraphics[width=3.8cm,height=3.8cm]{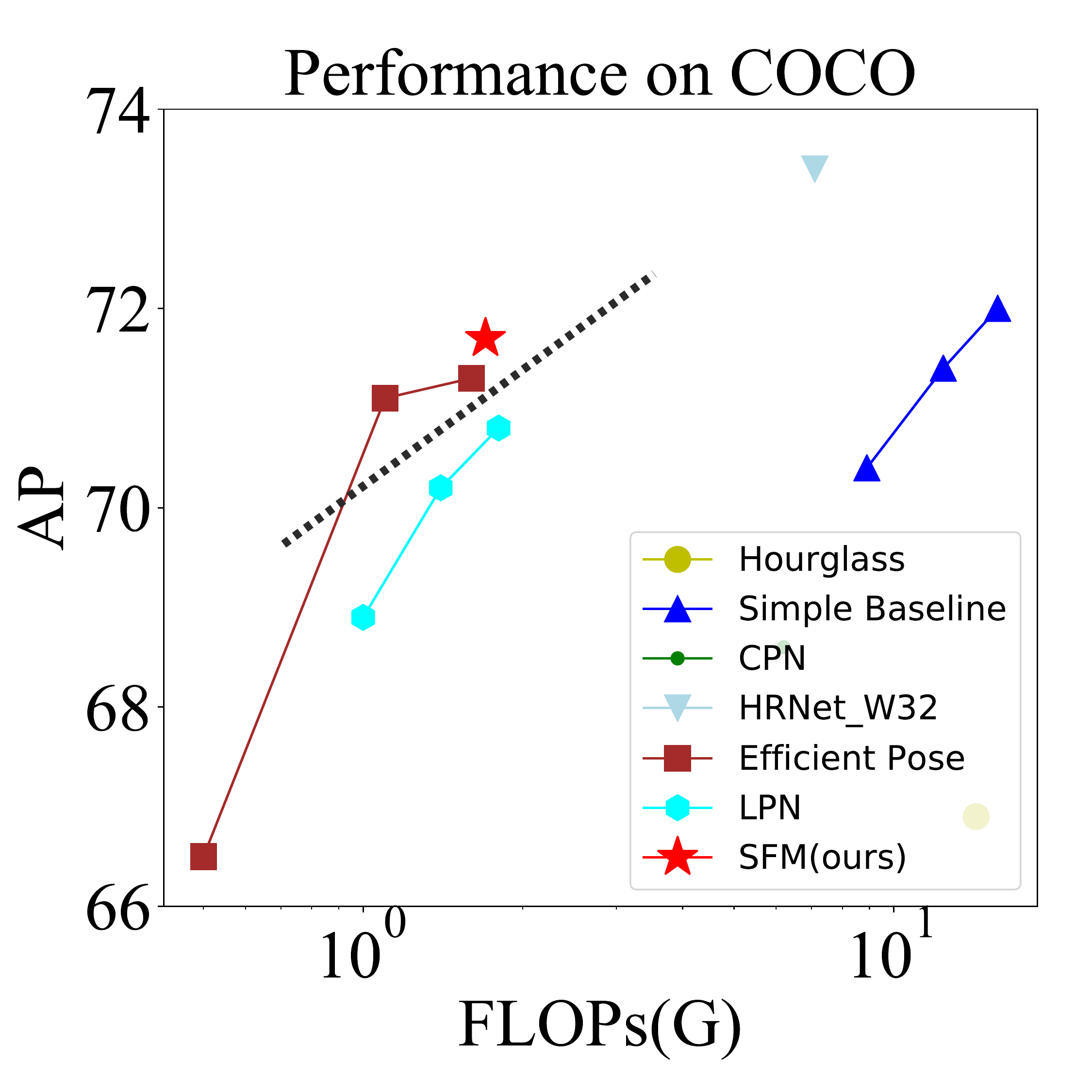}
}
\caption{Experiments on MPII validation and COCO validation. The closer to the top left corner the better the effect.}
\label{fig:firstpic}
\end{figure}

In order to have these two characteristics in networks, existing methods are basically based on feature pyramid network~\cite{yang2017learning, lin2017feature} such as U-net~\cite{ronneberger2015u} and Hourglass~\cite{newell2016stacked} and utilize spatial attention block~\cite{wang2018parameter}. For the former solution, this architecture does learn multi-level feature but only gains low value of metrics such as PCK, AP and so on due to quantization errors accumulated during frequent upsampling and downsampling. But upsampling and downsampling are indispensable because upsampling can reconstruct the feature map size and downsampling can expand the field of perception. For the later one, it is theoretically possible  for spatial attention block to learn the correlation of the space very well. But the experimental results show that it does not learn a good feature map, which can not extract the important areas well.

\begin{figure*}[h]
\centering
\includegraphics[scale=0.5]{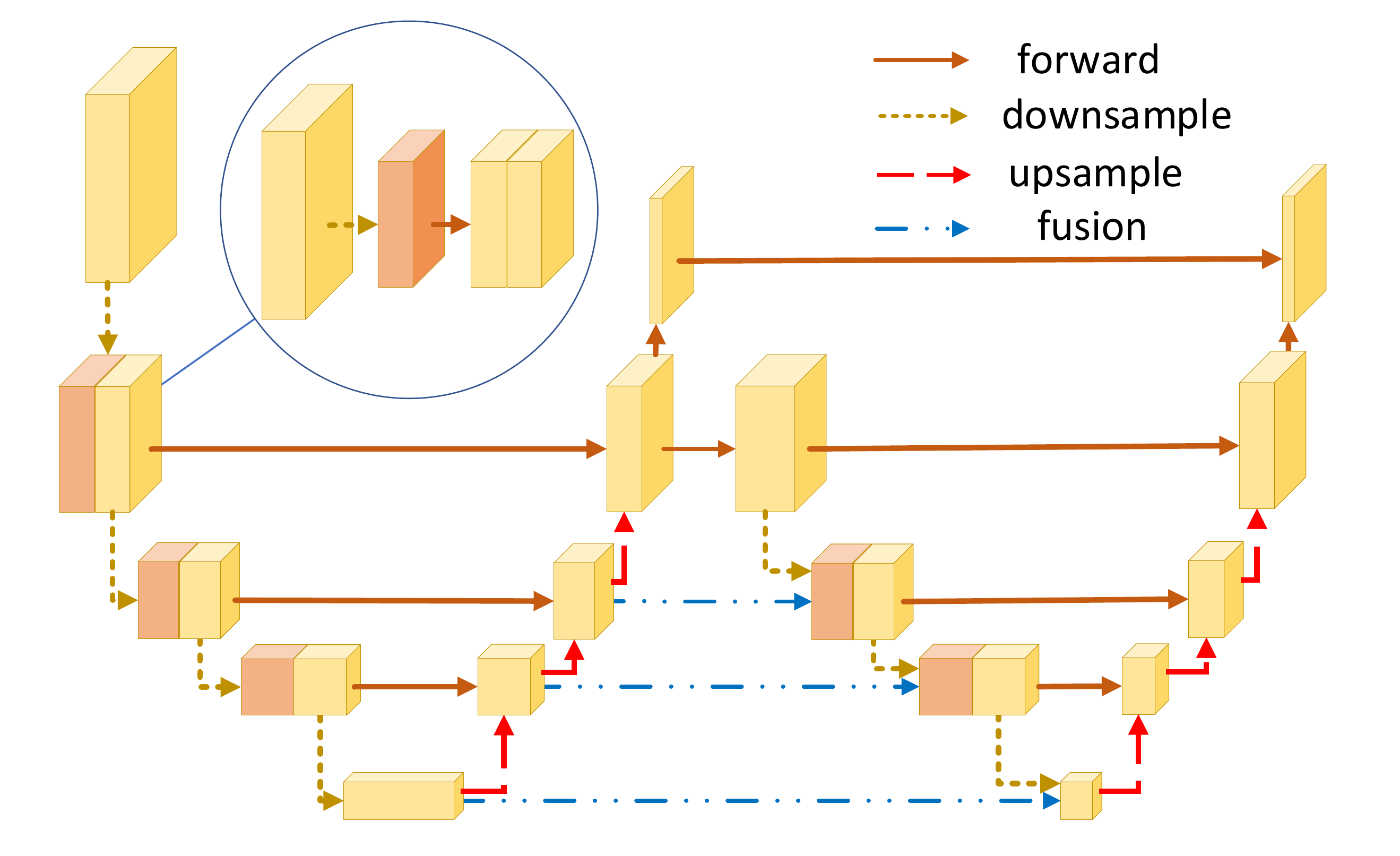}
\caption{The overview of SFM. Input $\in R^{3\times 256\times 256}$. Downsample uses convolution with stride 2. Upsample uses nearest neighbor interpolation. Forward operation will extract feature without changing size. Fusion operation will help combine two different feature maps. Output $\in R^{joints\times 128\times 128}$.}
\label{fig:overview}
\end{figure*}

To make some improvements in these methods, we propose a new architecture---SFM and HSA to solve aforementioned issues. We first expand the general pyramid network to a cascaded pyramid network to extend the network depth so as to increase the fault tolerance of the model. Furthermore, We make full use of the layers in the network and add fusion connections for different scales as much as possible. And we design a new attention mechanism called HSA, whose core is to compute the attention map of the attention map. We hope to gain position-precious attention map to learn proper spatial dependency and we embed them into the SFM network. Based on the results of our experiment shown in Fig. \ref{fig:firstpic}, these operations can largely improve the performance of network with few additional computation.

% To solve this problem, we first expand the general U-shaped network to a W-shaped network to extend the network depth in the hope of increasing the fault tolerance of the model. Further, We make full use of the layers in the network and add fusion connections for different scales as much as possible, which reaches an excellent performance in the experiment. 

% The existing posture network is basically based on the regression of heatmap~ and looks like shape U. But we find that a single U-like network only gain a small value of PCKh@0.1. In other words, quantization errors are accumulated during frequent upsample and downsample period. This motivates us to think: we can reduce this loss by enhancing the fusion of different level features. In the meantime, we realize that existing attention module is hard to learn during the training process. Based on the two serious problems above, network cannot achieve better performance in the task.

% We propose a new architecture-SFM and HSA to solve aforementioned issues. SFM base on two cascaded UNet-like model, between which important fusion bridges are constructed in order to help network layer save more context information or spatial information. HSA module is an attention mechanism, whose core is a two-order attention function. We find that without special handling HSA module can only be inserted into the output layer, so a residual structure is added to make the module more robust. From our experiment, these operations can largely inprove the performance of network with few additional computation.

Our contributions are:

1. We propose a novel network architecture SFM, which fuses more rich-scale information on the pyramid network by adding some important bridges between two pyramid framework. The performance improves dramatically, especially PCKh@0.1, meaning that the quantization errors are restrained well.

2. We propose a lighter attention block HSA, which apply a two-order attention operation on the image. This allows the feature map to finally generalize the global features to highlight important areas.

3. Our proposed SFM achieves 89.0 in PCKh@0.5, 42.0 in PCKh@0.1 on MPII validation dataset and 71.7 in AP, 90.7 in AP@0.5 on COCO validation dataset with only 1.7G FLOPs and 1.5M parameters.

%% file: related_work.tex
\section{Related Work}

\subsection{Pyramid Fusion}

Image fusion technology is the synthesis of two or more image information in the same scene under different spectral and spatial detail~\cite{chen2016fusion,bilodeau2011visible}. Pyramid fusion is an important approach in image fusion. Burt et al. proposed the algorithm in 1983 firstly~\cite{feng2020cpfnet,do2003framing}. It helps in tasks that have much multi-scale information, and is wildly used in human pose estimation. Each layer of the pyramid contains different frequency bands of the image. The fusion process is carried out on each layer separately. We can use different fusion operators on different layers depending on their features and details. 

\subsection{Cascaded Pyramid Network} 

It was firstly proposed in~\cite{ chen2018cascaded} for Multi-person Pose Estimation~\cite{cai2020learning}. It's a pyramid series connection model, which can consider the local information and global information of human joint points at the same time, leading to improve results. CPN consists of two parts: GlobalNet and RefineNet. Globalnet extracts the key points roughly, which can locate simple key points, such as eyes and hands. On the other hand, RefineNet processes the network which is difficult to identify. General cascaded pyramid frameworks~\cite{ chen2018cascaded} could combine different level feature maps, which reach an experimental good performance. However, it does not take full advantage of this design.

\subsection{Spatial Attention} 

Spatial attention~\cite{tootell1998retinotopy} has been widely used in high level vision tasks, such as image classification~\cite{wang2017residual}, image captioning~\cite{chen2017sca}, and visual question answering ~\cite{schwartz2017high,yu2017multi}. Attention module has the advantage to capture long-distance feature which is very important for human pose estimation task. Because some keypoints' positions deeply depend on other keypoints' information. For example, although the details of the crotch are completely covered by the clothes, the position can be roughly judged by the person's right shoulder and right knee. So spatial attention should works a lot in location-sensitive human pose estimation task. But in fact its attention map is hard to learn during the training process, which is verfied by the experiment part.

%% file: method.tex
\section{Method}

In this section, we firstly introduce the most important part of our method---SFM architecture. This network is full of fusion operation and utilizes only a small amount of computing resources.

And then we will go to the details of HSA. The block is so lightweight that can be inserted to any part of a network to learn long-range dependency.

\subsection{SFM}

 To further explore the potential of cascaded pyramid framework, we suppose different independent pyramid networks could have connection and interaction, by which the layer-wise aggregated features can contain more effective information and have more complex representing capacity. Towards this motivation, we come to the idea that build the bridges between two pyramid network to learn multi-level feature. The architecture is shown in Fig. \ref{fig:overview}.

Specifically, the key part of SFM---fusion can be mathematically written as
\begin{equation}
X_{l+1} = F(X_l, X_{history}),
\end{equation}
where $X_l$ denotes the feature map from former layer $l$, $X_{history}$ denotes the feature map from history, sharing the same scale with $X_{l+1}$. $X_{l+1}$ is the result of the fusion function $F(\cdot)$. $F(\cdot)$ can be flexibly selected, and for simplicity here we only consider $F(\cdot)$ in the form of add operation.

\begin{figure}[h]
\centering
\includegraphics[scale=0.3]{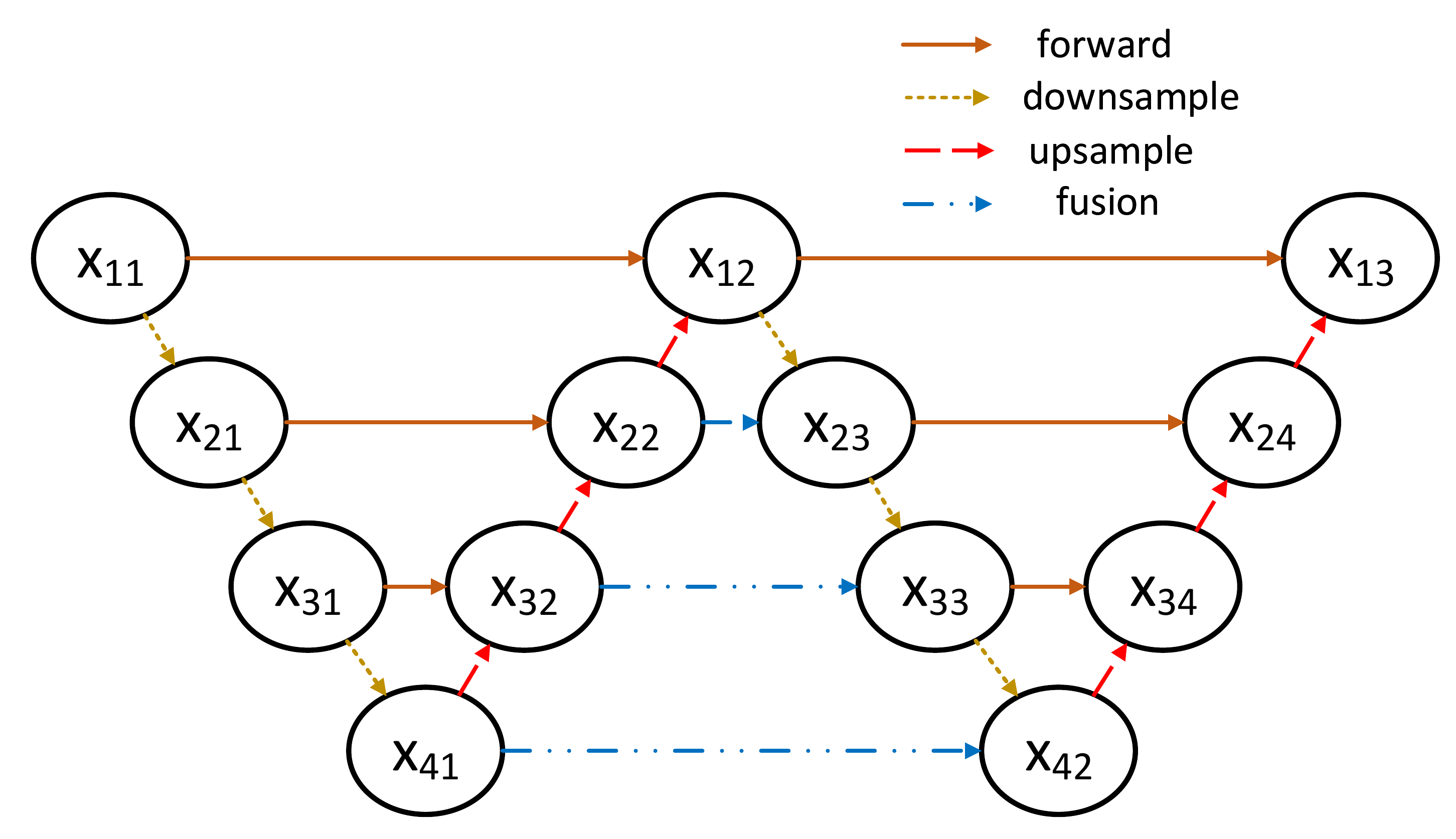}
\caption{ A simplified version of our proposed SFM.}
\label{fig:simple}
\end{figure}

Moreover, we add a connection from the middle layer to the final layer, which aims to mitigate the impact of too much upsampling on the accuracy of the final layer. Moreover, in the first pyramid network, we used the concat operation in the fusion operation, but in the second pyramid network, we used the add operation in the fusion operation. In this way we could preserve the information of the previous layer completely by concat operation and reduce some computation by add.

Let us present an example to show how the fusion operation can help learn multi-level feature. As shown in Fig. \ref{fig:simple}, supposing there are no bridges between $(X_{32}, X_{33})$, $X_{33}$ have to integrate the information from $X_{32}$ from $(X_{32}\rightarrow X_{22}\rightarrow X_{12}\rightarrow X_{23}\rightarrow X_{33})$. More importantly, the original information in $X_{32}$ will fade away with forward propagation. But with the help of the bridges$(X_{32}\rightarrow X_{33})$, $X_{33} $ can obtain the information from $X_{32}$ through this bridge, which is much shorter than the former one.

\subsection{HSA}

Although spatial attention has long been considered to be crucial for pose estimation due to its capability of capture long-distance features, the attention map is found to be difficult to learn during the training process.

\begin{figure}[htbp]
\centering
\subfigure[Spatial attention]{
\begin{minipage}[t]{0.5\linewidth}
\centering
\includegraphics[width=3.5cm]{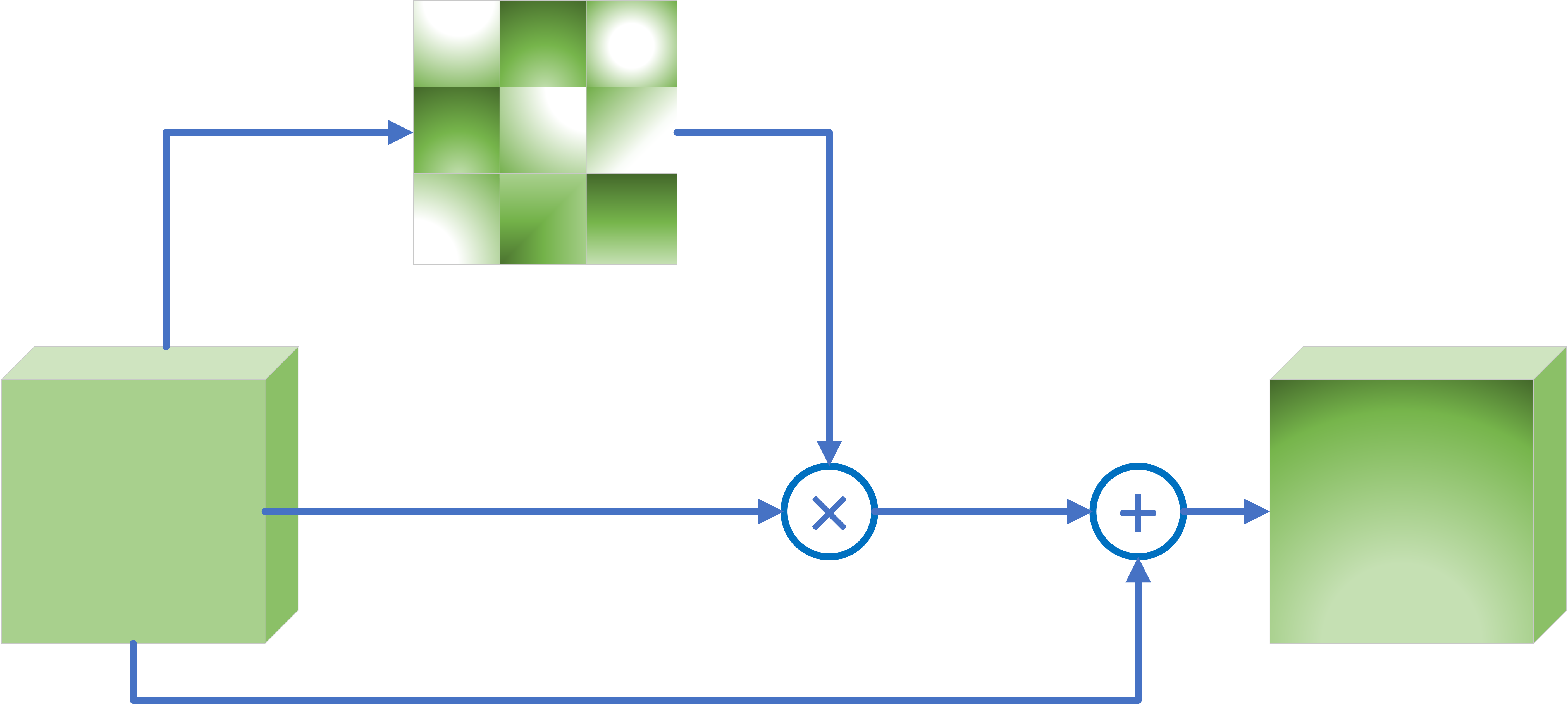}
\end{minipage}%
}%
\subfigure[High-level spatial attention(HSA)]{
\begin{minipage}[t]{0.5\linewidth}
\centering
\includegraphics[width=3.5cm]{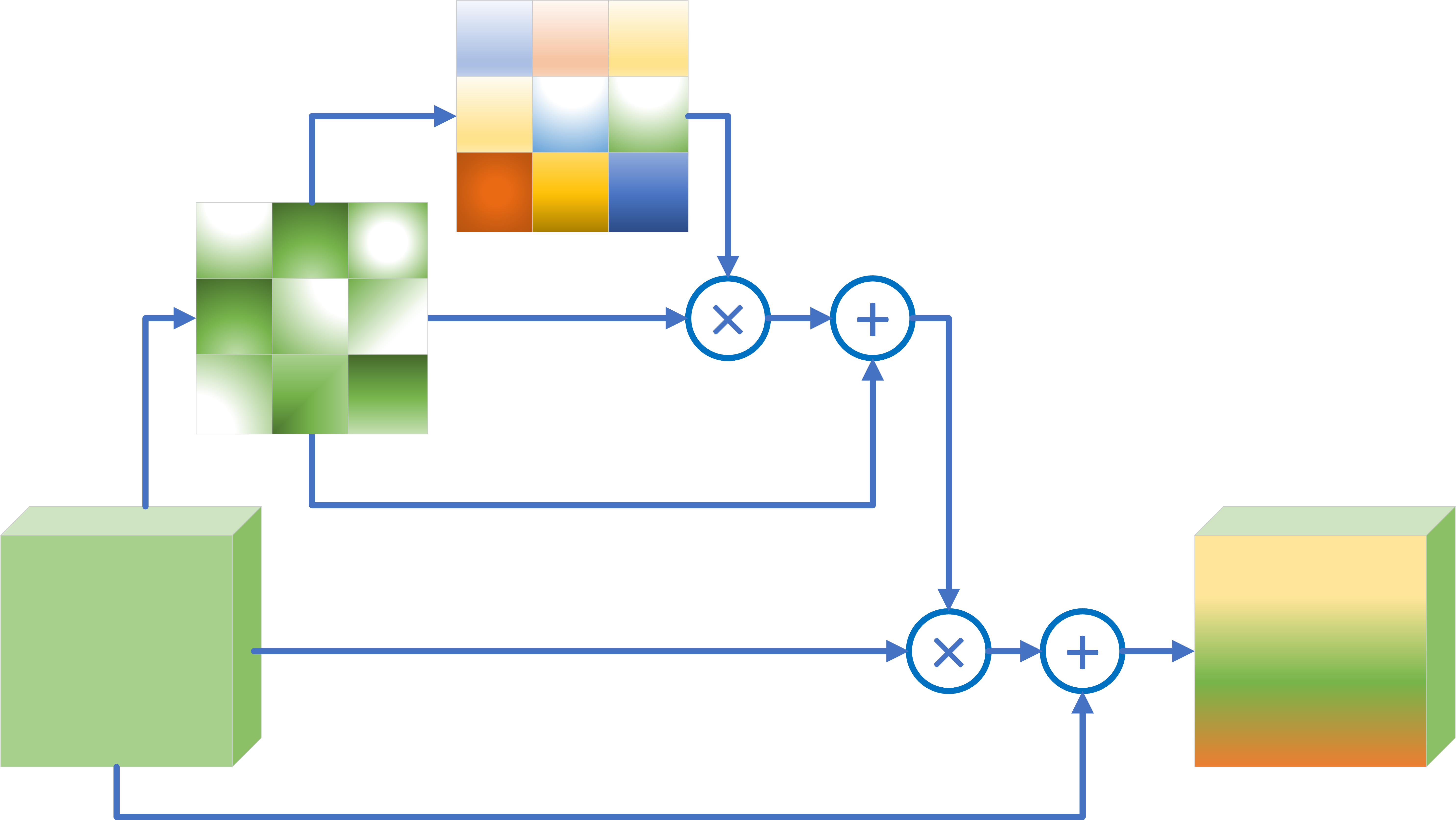}
\end{minipage}%
}%
\centering
\caption{Comparison between spatial attention and HSA.}
\label{fig:attention com}
\end{figure}

To find a way getting learnable attention map, we propose HSA which can help adjust the spatial attention map in good direction. Its core idea can be understood as computing the attention map of the attention map to get more precise attention map. As shown in Fig. \ref{fig:attention com}, we firstly extract an attention map denoted as $M$ from input feature map denoted as $f_{in}$. Then the attention map of $M$ is denoted as $M_{deeper}$. $\theta(\cdot)$ is the function to compute attention map of feature map. We then have,
\begin{equation}
M = \theta(f_{in}),
\end{equation}
\begin{equation}
M_{deeper} = \theta(M),
\end{equation}
\begin{equation}
f_{out}=f_{in} + f_{in}\cdot [M+M\cdot M_{deeper}].
\end{equation}
Supposing input feature map $f_{in}\in R^{C\times W\times H}$, $\theta(\cdot)$ function transforms it to low-level attention map $\theta(f_{in})=M \in R^{1\times W\times H}$. Then taking $M$ as the input feature map, we have high-level attention map  $\theta(M)=M_{deeper}\in R^{1\times W\times H}$. By doing product and addition operation twice, we have $f_{out}\in R^{C\times W\times H}$.

% \begin{figure}[htbp]
% \centering
% \begin{minipage}[t]{0.48\textwidth}
% \centering
% \includegraphics[width=6cm]{figure/spatial_attention.png}
% \caption{Spatial attention}
% \end{minipage}

% \begin{minipage}[t]{0.48\textwidth}
% \centering
% \includegraphics[width=6cm]{figure/High_Level_Attention.png}
% \caption{High-level spatial attention(HSA)}
% \end{minipage}
% \end{figure}

To ensure the stability of the network we add a shortcut, which means the network can propagate the original feature map forward even if attention does not learn anything. Meanwhile, we use the tanh activation function to guarantee the feature map zero-mean property.

In summary, the HSA module improves the accuracy of the entire network for long-range feature capture while consuming essentially no computational resources. In fact, it is not only very suitable for pose estimation tasks, but also very suitable for other position-sensitive tasks.

%% file: experiment.tex
\section{Experiments}

\subsection{MPII~\cite{andriluka20142d}}

\textbf{Dataset.} We train and evaluate our models on MPII. We use the standard train/val/test/ data split (26K for training, 3K for validation, and 11K for testing). The results are evaluated by Percentage of Corrected Keypoints (PCK), which indicates the fraction of correct predictions within an error threshold $\tau$ ($\tau$ = 0.5 for MPII, i.e. PCKh@0.5). 

\textbf{Training.}  We randomly rotate, flip and scale human boxes to augment training instances. The rotation range is $[-30^{\circ}, 30^{\circ}]$, and the scaling range is $[0.75,1.25]$. Half-body transform is also performed for data augmentation. The human boxes will be resized to $256\times256$. We use Adam as our optimizer. The initial learning rate is set as $1\times10^{-3}$ and it decays at epoch 180 and epoch 260. The batch size is 32 and we train all models by 300 iterations on 2 RTX3090 GPUs for around 8 hours.

\textbf{Testing.} During testing, using the same strategy in the training section, images and their flipped versions are all evaluated. The predicted heatmaps~\cite{bulat2016human} are averaged and then blurred by a Gaussian filter. To obtain the final location, instead of directly extracting the location of the highest response in heatmaps, we adjust it by a quarter offset in the direction of the second-highest response.

% \begin{figure*}[h]
% \centering
% \includegraphics[scale=0.5]{figure/Comparison.png}
% \caption{Comparison.}
% \label{1}
% \end{figure*}

\textbf{Comparison with Existing Methods.} Intensive experiments are done on SFM. Reference models mainly include the families of  CPMs~\cite{wei2016convolutional}, DLCM~\cite{ai2018learning}, Simple-Baseline~\cite{xiao2018simple}, OpenPose~\cite{cao2019openpose} and EfficientPose~\cite{groos2021efficientpose}. The results on MPII validation are in Table \ref{tag:MPII}. Our SFM has only 1.7GFLOPs and 1.5Params but achieves comparable results with large models and efficient models at PCKh@0.5 and achieves significant advantage at PCKh@0.1. This is because we use a $128*128$ heatmap output instead of wasting computation on stacking of layers. At the same time, HSA allows it to learn the precise information better and reduce the error of decode operation.
\begin{table}[htbp]
\centering
\caption{COMPARISONS OF RESULTS ON MPII VALIDATION}
\begin{tabular}{|p{1.8cm}|p{1.4cm}|p{0.9cm}|p{1.2cm}|p{1.2cm}|}%l=left, r=right,c=center分别代表左对齐，右对齐和居中，字母的个数代表列数
\hline
Method & Params(M) & GFLOPs & PCKh@0.5 & PCKh@0.1 \\ \hline
CPMs& 31.0& 175.0& 88.0& - \\ \hline
DLCM& 34.0& 12.0& 88.5& - \\ \hline
SimpleBaseline& 34.0& 12.0& 88.5& - \\ \hline
OpenPose& 25.94& 160.36& 87.60& 22.76 \\ \hline
EfficientPoseRT& 0.46& 0.87& 82.88& 22.76 \\ \hline
EfficientPose \romannumeral1& 0.72& 1.67& 85.18& 26.49 \\ \hline
EfficientPose \romannumeral2& 1.73& 7.70& 85.18& 30.17 \\ \hline
EfficientPose \romannumeral3& 3.23& 23.35& 89.51& 30.90 \\ \hline
EfficientPose \romannumeral4& 6.56& 72.89& 89.75& 35.63 \\ \hline
\textbf{SFM}& \textbf{1.50}& \textbf{1.70}& \textbf{89.0}& \textbf{42.0} \\ \hline
\end{tabular}
\label{tag:MPII}
\end{table}

\textbf{Ablation Study.} We investigate each component, specifically SFM and HSA.  We respectively remove all bridges between two pyramid networks and HSA module as comparison experiments. The results on MPII validation are shown in Table \ref{tag:xiaorong}. SFM can improve PCKh@0.5 score by 1.7 and PCKh@0.1 score by 0.7. HSA can improve PCKh@0.5 score by 0.6 and PCKh@0.1 score by 0.8. And almost no extra flops and parameters are introduced.

\begin{table}[htbp]
\centering
\caption{ABLATION STUDY}
\begin{tabular}{|p{2.4cm}|p{1.3cm}|p{0.75cm}|p{1.2cm}|p{1.1cm}|}
\hline
Method & Params(M) & GFLOPs & PCKh@0.5 & PCKh@0.1 \\ \hline
Cascaded Pyramid& 1.5& 1.7& 86.3& 39.5 \\ \hline
SFM (without HSA)& 1.5& 1.7& 88.0& 41.2 \\ \hline
SFM (with HSA)& 1.5& 1.7& \textbf{89.0}& \textbf{42.0} \\ \hline
\end{tabular}
\label{tag:xiaorong}
\end{table}

\begin{table}[h]
    \centering
    \caption{COMPARISONS OF RESULTS ON COCO VALIDATION} 
    \begin{tabular}{|c|c|c|c|c|}
    \hline
         Method             &Pretrain(\%)  &Params(M)(\%)  &GFLOPs(\%)  &AP(\%)         \\ \hline
         Hourglass          &N             &25.1           &14.3        &66.9           \\ \hline
         CPN                &Y             &102.0          &6.2         &68.6           \\ \hline
         LPN-50             &N             &2.9            &1.0         &68.9          \\ \hline
         SBline-R50         &Y             &34.0           &8.9         &70.4           \\ \hline
         HRNet-W32          &N             &28.5           &7.1         &73.4           \\ \hline
         EfficientPose-A    &N             &1.3            &0.5         &66.5           \\ \hline
         EfficientPose-B    &N             &3.3            &1.1         &71.1           \\ \hline
         EfficientPose-C    &N             &5.0            &1.6         &71.3           \\ \hline
         \textbf{SFM}       &N             &\textbf{1.5}   &\textbf{1.7}&\textbf{71.7}  \\ \hline
    \end{tabular}
    \label{tag:CICIDS2017}
\end{table}

\subsection{COCO~\cite{kim2019evaluation}}
\textbf{Dataset.} We use COCO train2017 as our training dataset, which includes 57K images and 150K person instances. We use COCO val2017 (5K images) to evaluate our models. The evaluation metric is based on Average Precision (AP).

\textbf{Training\& Testing.} This section is same as section $\mathrm{IV}$.A and $\mathrm{IV}$.B

\textbf{Comparison with Existing Methods.} Intensive experiments are done on SFM. Reference models mainly include the families of Hourglass~\cite{newell2016stacked}, CPN~\cite{chen2018cascaded}, LPN~\cite{zhang2019simple}, HRNet~\cite{sun2019high}, EfficientPose-NAS(note: this paper is different from the one in MPII )~\cite{zhang2020efficientpose}. The results on COCO validation are in Table \ref{tag:CICIDS2017} ,Our SFM has only 1.7GFLOPs and 1.5Params but achieves comparable results with large models and efficient models at AP.

\setcounter{figure}{5}

\begin{figure*}[t]
    \centering
    \begin{minipage}[t]{0.95\textwidth}
        \centering
        \includegraphics[width=\columnwidth]{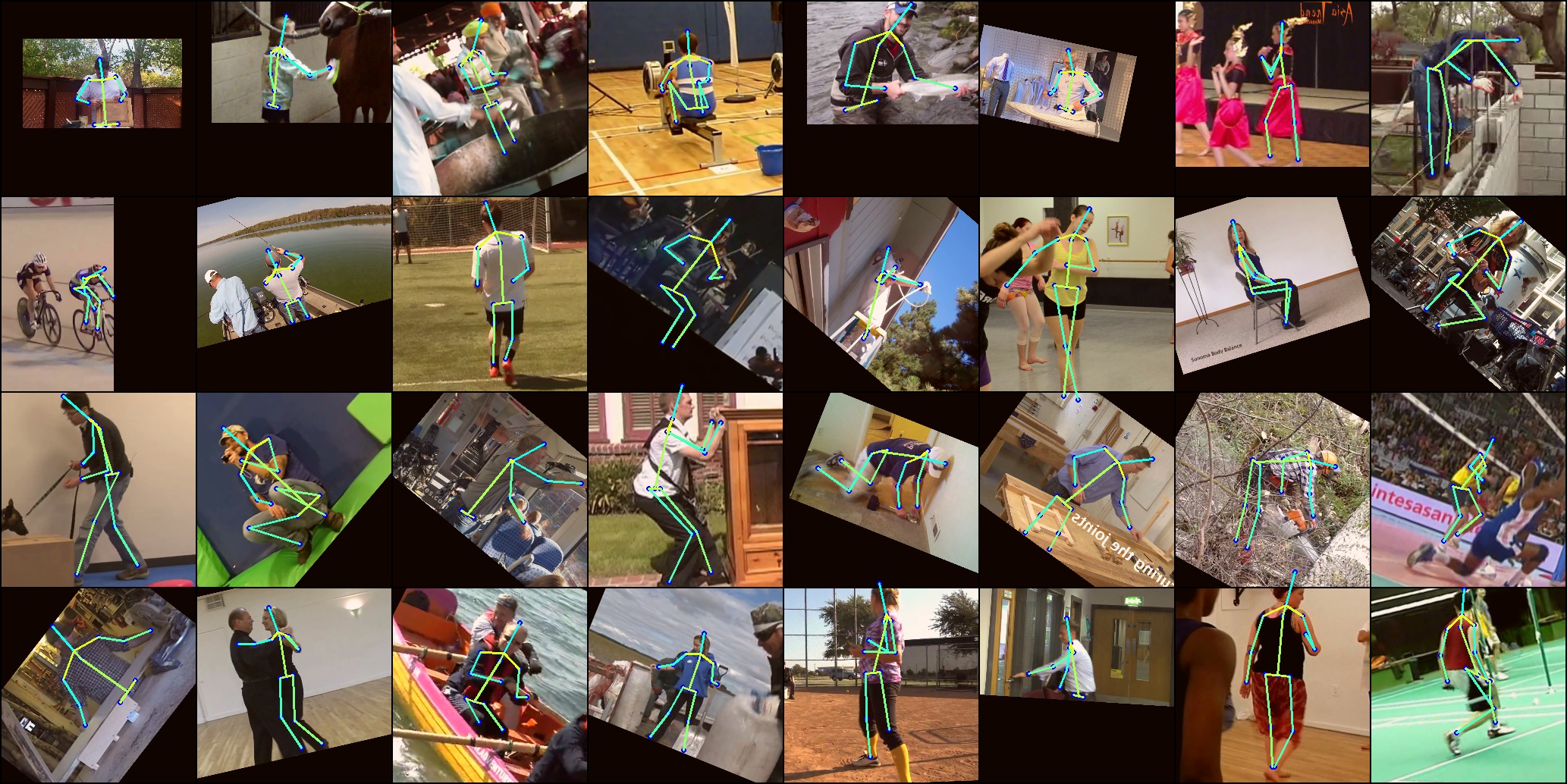}
        \caption{The visualization result of our model on MPII validation dataset.}
        \label{fig:results}
    \end{minipage}
\end{figure*}

\setcounter{figure}{4}

\begin{figure}[H]
    \centering
    \subfigure[original image]{
        \includegraphics[width=7cm]{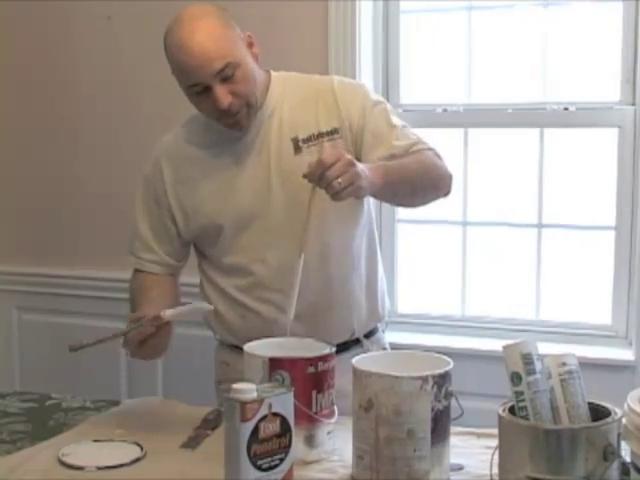}
        }
    \quad
    \subfigure[attention map in HSA]{
        \includegraphics[width=3.5cm]{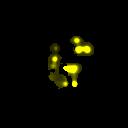}
        }
    \quad
    \subfigure[attention map in general spatial attention]{
        \includegraphics[width=3.5cm]{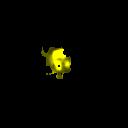}
        }
    \caption{Visualization comparison of the attention map of HSA and that of general spatial attention map}
    \label{attention—vis}
\end{figure}

\setcounter{figure}{6}

\subsection{Visualization}

\textbf{Attention Map Visualization.} Fig. \ref{attention—vis} shows that the comparison between the attention map in HSA and the attention map in general spatial attention. We can observe that many keypoints are explicitly highlighted in the attention map of the HSA, which confirms attention map in HSA can learn more correlation between keypoints than that in general spatial attention. 

\textbf{Predicted Keypoints Visualization.} Fig. \ref{fig:results} shows that the performance of our model on the MPII validation is excellent.

\subsection{Discussions}

\textbf{Further Discussion on SFM.}  Multi-level feature fusion is a very important operation for most computer vision tasks. But SFM is in fact an artificially designed network, which cannot guarantee that this is the best structure for feature fusion. So in the future, we want to explore out a new backbone for learning which layer to do fusion operation. And fusion operation is also important in other tasks, so we will do extra experiments on image classification, object detection, semantic segmentation.etc.

\textbf{Further Discussion on HSA.} High-level spatial attention has been experimentally proven to be very effective. But it must be in a specific location to achieve such effect. In our experiment, we embed four modules into two deepest positions of the downsampling and two output layers. In other conditions, effect is not as good as former experiment. We try to come up with some explanations. In the deepest downsampling position, the feature map becomes particularly small in size after a series of downsampling, and the need for spatial sensitivity is very high, so the results are much better when adjusted using HSA. In the final layer position, the feature map is about to be output as the final result, so the information of the whole map is integrated and corrected, and will not be corrupted by the convolution with small feeling fields again.

%% file: conclusion.tex
\section{Conclusion}

Most existing human pose estimation models use pyramid framework to extract multi-level feature  and lack the capacity to integrate long-distance feature. In this paper, we propose a new architecture---SFM which could extract richer multi-level feature and a new attention module---HSA to capture long-distance dependency. With the help of SFM and HSA, our network achieve comparable or even better accuracy with much smaller model complexity. Specifically, our SFM achieve 89.0 in PCKh@0.5, 42.0 in PCKh@0.1 on MPII validation dataset and 71.7 in AP, 90.7 in AP@0.5 on COCO validation dataset with only 1.7G FLOPs and 1.5M parameters.

\vfill